# CephGPT-4: An Interactive Multimodal Cephalometric Measurement and Diagnostic System with Visual Large Language Model

Lei Ma[1]* Jincong Han[1], Zhaoxin Wang[1], Dian Zhang[1]

1. School of Information Science and Technology, University of Nantong; Nantong, China, CN 226019

    *Correspondence: malei@ntu.edu.cn

**Abstract:** Large-scale multimodal language models (LMMs) have achieved remarkable success in general domains. However, the exploration of diagnostic language models based on multimodal cephalometric medical data remains limited. In this paper, we propose a novel multimodal cephalometric analysis and diagnostic dialogue model. Firstly, a multimodal orthodontic medical dataset is constructed, comprising cephalometric images and doctor-patient dialogue data, with automatic analysis of cephalometric landmarks using U-net and generation of diagnostic reports. Then, the cephalometric dataset and generated diagnostic reports are separately fine-tuned on Minigpt-4 and VisualGLM. Results demonstrate that the CephGPT-4 model exhibits excellent performance and has the potential to revolutionize orthodontic measurement and diagnostic applications. These innovations hold revolutionary application potential in the field of orthodontics.

## 1 Introduction

Malocclusion refers to the misalignment of teeth, abnormal dental arch relationships, and abnormal facial and jaw positioning caused by genetic or environmental factors. It is a common oral condition, with approximately 56% of teenagers worldwide experiencing malocclusion issues [1]. The health of the teeth and jaws not only directly affects facial aesthetics but also impacts jaw function and oral health. Patients require orthodontic treatment to restore the aesthetic appearance and normal functionality of the oral and jaw system.

Cephalometric analysis is a method used to analyze the relationships between the skull, teeth and other bone structures. It is widely used in orthodontic clinical diagnosis, treatment planning, treatment evaluation, craniofacial structure research, and assessment of childhood growth and development [2]. Traditional cephalometric analysis involves manual annotation of anatomical landmarks on lateral X-ray images of the head. Measurements and calculations are then performed on these landmarks, including angles and distances, to analyze and assess anomalies in the anatomical structures of the oral cavity and develop treatment plans.

However, traditional cephalometric analysis faces three major obstacles. Firstly, the calibration of anatomical landmarks requires manual annotation by clinically trained doctors, which is a time-consuming task prone to errors [3,4]. Secondly, there is an imbalance in medical resources, with overcrowding in major hospital dental departments. Patients tend to seek treatment in dental clinics for non-severe oral conditions. However, doctors and nurses in dental clinics may have limited knowledge and training, making accurate landmark annotation challenging. Lastly, providing accurate interpretation of cephalometric analysis results and treatment recommendations is a significant challenge. Thus, there is an urgent need for a large-scale orthodontic cephalometric diagnostic model to address these issues.

In recent years, large language models (LLMs), represented by ChatGPT, have demonstrated remarkable performance and achieved significant success in various fields. LLMs refer to complex models trained on a massive amount of data and powerful computing resources in the field of natural language processing (NLP). These models aim to simulate human language understanding and generation abilities by learning the structure, semantics, and context of language through extensive textual data.

With the emergence of more and more large language models, innovative developments in multimodal models have been promoted. GPT-4, in particular, exhibits extraordinary performance compared to other large language models. However, detailed information about its training strategies and weight parameters is not publicly available, making it challenging for researchers to further train and fine-tune it. On the other hand, open-source large language models such as MiniGPT-4 and VisualGLM have demonstrated impressive performance in general domains, but their understanding capabilities may be insufficient in specific fields such as healthcare, hindering them from fully showcasing their true potential. Addressing the limitations of MiniGPT-4 and VisualGLM in the field of orthodontic medicine, this paper proposes the CephGPT-4 model, the world's first Interactive Multimodal Cephalometric Measurement and Diagnostic System with Visual Large Language Model (Fig.1).

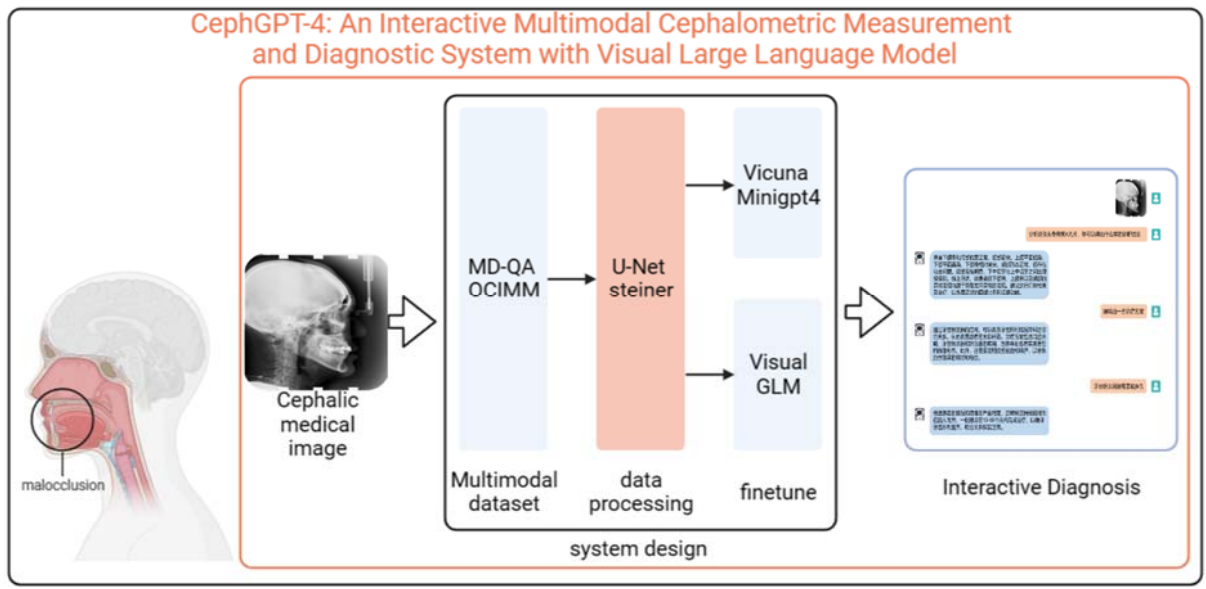

Figure 1. Illustration of CephGPT-4. CephGPT-4 incorporates a fine-tuned version of MiniGPT-4 and VisualGLM on a vast collection of cephalometric images and doctor-patient dialogue data. With CephGPT-4, users could upload lateral X-ray images of the head are annotated with anatomical landmarks, which are used to perform measurements and calculations. These measurements include angles and distances, and they are utilized to analyze and evaluate abnormalities in the oral cavity's anatomical structures and provided treatment plans based on this analysis.

## 2 Related Works

**2.1 Large Language Models**

Recently, researchers have discovered a model called LLaMa, which was proposed in 2021. It not only provides multiple models with parameter sizes ranging from 7 billion to 65 billion, along with detailed descriptions of their architectures and training strategies but also offers performance evaluations on various tasks [5]. Alpaca, a model based on LLaMa, demonstrates excellent performance and application potential in programming tasks through guided fine-tuning [6]. To enhance the functionality of Chinese language models, Du et al. developed a Chinese language model with 130 billion parameters using the GLM architecture, trained and fine-tuned it [7]. ChatGLM, which aligns human intentions through supervised fine-tuning, demonstrates its potential in the Chinese context [8].

## 2.2 Pre-trained Models in Biomedical Domain

In the field of biomedical natural language processing (NLP), several large language models derived from ChatGPT have been introduced, including ChatDoctor, Med-Alpaca, PMC-LLaMA, DoctorGLM, and Huatuo. These models are all based on the OphGLM model, which performs disease assessment and diagnosis using fundus images and incorporates ophthalmic knowledge data and real medical conversations. Furthermore, the OphGLM model integrates visual capabilities and establishes a new dataset for ophthalmic multimodal instruction following and dialogue fine-tuning. Experimental results demonstrate the outstanding performance of the OphGLM model, indicating its potential for revolutionary changes in ophthalmic clinical applications [10]. LLaVA-Med is a visual language dialogue assistant that can answer open-ended biomedical research questions related to biomedical images. By generating open-ended instruction-following data from a large-scale and widely covered biomedical image captioning dataset and applying a novel curriculum learning approach for fine-tuning a large-scale general-domain visual language model, LLaVA-Med exhibits excellent multimodal dialogue capabilities and can answer queries regarding biomedical images based on open-ended instructions. The fine-tuning results of LLaVA-Med outperform previous supervised state-of-the-art methods on three standard biomedical visual question-answering datasets [11]. HuatuoGPT, during supervised fine-tuning, utilizes both compressed data from ChatGPT and real-world data from doctors. By training a reward model to align the language model with the strengths of both data sources, HuatuoGPT employs reinforcement learning from AI feedback (RLAIF) to enhance its performance. To evaluate and benchmark the model, a comprehensive evaluation scheme comprising automatic and manual evaluation metrics is proposed. Experimental results show that HuatuoGPT achieves state-of-the-art performance among open-source LLM models in medical consultation, particularly in GPT-4 evaluation, human evaluation, and medical benchmark datasets. Notably, the compressed language model (i.e., HuatuoGPT) outperforms its teacher model, ChatGPT, in most cases through the use of additional real-world data and RLAIF [11].

XrayGPT is a novel dialogue-based medical visual language model designed to analyze and answer open-ended questions about chest X-rays. The model aligns a medical visual encoder (MedClip) with a fine-tuned large-scale language model (Vicuna) using a simple linear transformation. This alignment enables the model to possess excellent visual dialogue capabilities and a deep understanding of X-rays and medical domain knowledge. To improve the language model's performance in a medical context, the authors generated 217,000 interactive and high-quality summaries from free-text radiology reports to enhance the language model's performance through the fine-tuning process. This approach opens up new avenues for advancing the automated analysis of chest X-rays [12].

PMC-LLaMA is an open-source language model that has been fine-tuned on a total of 4.8 million biomedical academic papers to inject medical knowledge and enhance its capabilities in the medical field. The model has undergone preliminary evaluations on three biomedical question-answering datasets, including PubMedQA, MedMCQA, and USMLE. The results show improved performance in understanding domain-specific concepts in the biomedical field and achieve high performance on question-answering benchmarks after fine-tuning [13].

PMC-VQA is a visual instruction fine-tuning model for medical visual question-answering (MedVQA). Efficient interpretation of critical clinical information in medical images is crucial, which is why the focus is on addressing MedVQA. First, the MedVQA questions are restructured into a generation task, which naturally follows the human-machine interaction process. A generation-based model is proposed by aligning the visual information from a pre-trained visual encoder with a large-scale language model, enabling medical visual understanding. Secondly, an extensive process is established to construct a large-scale MedVQA dataset called PMC-VQA, comprising 227,000 VQA pairs and 149,000 images, covering various modalities and diseases. The proposed model is pre-trained on PMC-VQA and fine-tuned on multiple public benchmark datasets such as VQA-RAD and SLAKE, significantly outperforming existing works. Additionally, a manually verified test set with higher challenge is introduced to evaluate the model's performance, which even the best models find difficult to solve [14].

## 3. methods

**3.1 Dataset Construction**

In order to improve the question-answering performance of VisualGLM and Minigpt4 in the orthodontic diagnosis dialogue task, we constructed a multimodal orthodontic medical dataset that includes doctor-patient dialogue data, orthodontic medical images, and orthodontic analysis reports.

3.1.1 Dialogue Dataset

First, we created the Medical Domain Question-Answering Dataset (MD-QA). This dataset consists of 59,642 doctor-patient dialogue records in both English and Chinese (29,624 in English and 29,618 in Chinese). The dialogues cover topics such as the interpretation of orthodontic images, diagnosis, and treatment recommendations.

3.1.2 Multimodal Orthodontic Craniofacial Imaging and Measurement Dataset

Next, we collected real clinical case samples and established the Orthodontic Craniofacial Imaging and Measurement Multimodal Dataset (OCIMM Dataset), which covers various types and conditions of orthodontic craniofacial imaging and measurement. The OCIMM Dataset includes orthodontic medical images and corresponding textual diagnostic reports.

The orthodontic medical images were obtained from relevant databases and image resources in the field and were annotated by professional doctors [15]. The orthodontic analysis reports were generated using an improved U-Net algorithm and the Steiner analysis method.

We designed a diagnostic neural network model for orthodontic measurements based on the U-Net architecture, which includes both local and global network models. The local network adopts a general U-Net to learn local features from multiple domains, while the global network consists of parallel dilated convolutions to extract global features and address the problem of landmark point ambiguity. The method takes lateral cephalograms (craniofacial measurement analysis) as the primary input and performs measurements and analysis on multiple anatomical landmarks of the head.

The orthodontic analysis reports were automatically generated based on the Steiner analysis method, using a programmed calculation of angles, distances, and relationships between the landmarks. The Steiner analysis method is a measurement and analysis method for craniofacial imaging that was proposed by Steiner in 1953. It consists of 14 measurement parameters and was selected from previously proposed analysis methods by Riede, Downs, and others [15]. The Steiner analysis method determines the positions of the skull, teeth, and other bones based on the angles, distances, and relationships between the landmarks. It analyzes and evaluates the anomalies in the anatomical structures of the oral cavity and assesses the patient's maxillofacial morphology.

3.1.3 Data Cleaning

Finally, we performed data cleaning on the MD-QA and OCIMM datasets, removing low-quality images and dialogue data.

**3.2 Model Fine-tuning**

In this study, we performed fine-tuning on both MiniGPT-4 and VisualGLM models using the MD-QA and OCIMM datasets to compare their performance in the cephalometric analysis and diagnosis dialogue task.

3.2.1 Fine-tuning on MiniGPT-4

MiniGPT-4 is an advanced fine-tuning model that combines the visual encoder BLIP-2 and the large language model Vicuna to provide novel visual language capabilities. In this process, the visual encoder and LLM (Vicuna) from BLIP-2 are frozen and aligned to jointly provide powerful functionalities.

First, we fine-tuned the Vicuna-7B model using a large amount of medical dialogue data from MD-QA to enable it with the ability to handle Chinese medical conversations and answer common medical questions. Then, we aligned Vicuna-7B with the visual encoder and performed fine-tuning on MiniGPT-4 using the OCIMM dataset. We employed pre-defined prompts and instruction sets to standardize the format of the training samples. This helps ensure that the model can correctly infer and answer questions given the provided instructions.

"###Doctor:<ImageFeature></Img><Instruction>###Assistant:". The <Instruction> part is divided into two sections. The first section randomly selects instructions from the pre-defined instruction set to ensure that the model can respond correctly to different forms of instructions. The instruction set includes various formats of instructions such as "What diagnosis can you provide based on this cephalometric X-ray image?", "Please analyze this cephalometric X-ray image from the perspective of an orthodontist", "Based on this cephalometric lateral X-ray, what diagnostic recommendations can you provide?", etc. The second section aims to enhance the model's understanding of cephalometric X-ray images by adding several measurements used during dataset creation. The prompt format is as follows（see Table 2）

Table 2 The format of Prompt

| Prompt（Chinese） | Prompt（Translated to English） |
| --- | --- |
| 在你作为一位口腔正畸医生的身份下，基于这张头骨侧视 X 光片，你能够提供哪些医学诊断信息? \n 参考指标:SNA 角:84.41,SNB 角:85.7,ANB 角:-1.29,Y 轴角:61.28,MP-FH 角:28.03,面 角:94.25,U1-NA 距离:6.34,L1-NB 距离:6.6,Po-NB 距离:0.08 | In your role as an orthodontist, based on this lateral cephalometric X-ray of the skull, what diagnostic information can you provide? \n Reference measurements: SNA angle: 84.41, SNB angle: 85.7, ANB angle: -1.29, Y-axis angle: 61.28, MP-FH angle: 28.03, facial angle: 94.25, U1-NA distance: 6.34, L1-NB distance: 6.6, Po-NB distance: 0.08 |
| 根据这个头骨侧面 X 光图片，给出你的专业诊断. \n 参考指标:SNA 角:84.54,SNB 角:84.27,ANB 角:0.27,Y 轴 角:58.64,MP-FH 角:26.6,面 角:89.64,U1-NA 距离:6.18,L1-NB 距离:8.91,Po-NB 距离:-3.56 | Based on this lateral cephalometric X-ray image of the skull, provide your professional diagnosis. \nReference measurements: SNA angle: 84.54, SNB angle: 84.27, ANB angle: 0.27, Y-axis angle: 58.64, MP-FH angle: 26.6, facial angle: 89.64, U1-NA distance: 6.18, L1-NB distance: 8.91, Po-NB distance: -3.56. |
| 这是一个头骨侧面 X 光照片，根据它你能够提供哪些诊断建议? \n 参考指标:SNA 角:80.01,SNB 角:73.87,ANB 角:6.14,Y 轴角:63.03,MP-FH 角:31.03,面 角:96.67,U1-NA 距离:0.41,L1-NB 距离:11.55,Po-NB 距离:-0.94 | This is a lateral cephalometric X-ray image of the skull. Based on it, what diagnostic suggestions can you provide? \nReference measurements: SNA angle: 80.01, SNB angle: 73.87, ANB angle: 6.14, Y-axis angle: 63.03, MP-FH angle: 31.03, facial angle: 96.67, U1-NA distance: 0.41, L1-NB distance: 11.55, Po-NB distance: -0.94. |

3.2.2 Fine-tuning on Visual GLM

In addition, we also performed fine-tuning on the VisualGLM-6B model, which is based on ChatGLM-6B as the base model and establishes a bridge between the visual and language models using VIT and Qformer.

However, due to the lack of domain-specific training data, its performance on specific tasks is limited. Therefore, fine-tuning VisualGLM can improve its performance in the cephalometric analysis and diagnosis dialogue task. We performed fine-tuning on VisualGLM using the OCIMM dataset. To retain its original multi-turn dialogue and image understanding capabilities, we employed an efficient fine-tuning method called Low-Rank Adaptation. During fine-tuning, we faced the challenge of a relatively small dataset size. To ensure that the model can learn new knowledge while preserving its multi-turn dialogue and image understanding capabilities, we reduced the number of trainable parameters and only performed low-rank adaptation on the 0th layer, thus retaining the model's multi-turndialogue and image understanding abilities. Similarly, we pre-defined an instruction set and randomly sampled prompts during fine-tuning to ensure that the model can respond correctly to different forms of instructions. Finally, we compared the response performance of the lora-fine-tuned Visual GLM model with the ptuning-fine-tuned one, demonstrating the effectiveness of the fine-tuning.

## 4. Results

In this paper, we introduced a novel cephalometrics language-visual fine-tuning model called CephGPT-4. The model was fine-tuned on MiniGPT-4 and VisualGLM, two state-of-the-art visual language models in the general domain.

The experimental results showed that fine-tuning VisualGLM significantly improved its performance in the specific domain of visual question answering. The fine-tuned model achieved better accuracy and relevance in answering questions compared to the baseline model, demonstrating higher performance. By analyzing the generated results, we observed that both models were able to accurately describe the relationship between the upper and lower jaws and assess the chin morphology. The MiniGPT-4 fine-tuned model provided more detailed and accurate diagnostic results regarding the sagittal plane angles and could accurately identify the malocclusion type, providing more detailed oral recommendations. On the other hand, the fine-tuned VisualGLM-6B model provided a more detailed identification of the relationship between the upper and lower central incisors and retained its multi-turn dialogue capability for further communication and inquiry.

Overall, CephGPT-4 demonstrated excellent visual transformation abilities and a deep understanding of cephalometric X-ray images. This work opens up new possibilities for advancing automated analysis of dental malocclusions and provides support for the development of large language models in the medical field. By introducing the cephalometrics language-visual fine-tuning model, we can expect more precise and efficient correction of dental malocclusions in the field of dentistry.

CephGPT-4 was trained on a large-scale dataset of cephalometric X-ray (cephalogram) images and related medical literature (doctor-patient dialogue dataset) to enable automated analysis and measurements of cephalometric X-ray images. By learning the correlations between the head structures and features, the model can accurately measure and diagnose skeletal structures, facial proportions, and other characteristics in the images.

## 5 Discussion

This study proposes a novel multimodal cephalometric measurement analysis and diagnostic dialogue model called CephGPT-4, which can automatically analyze cephalometric medical images and provide diagnostic results and treatment advice.

First, the study constructs a multimodal orthodontic medical dataset, including a large amount of cephalometric medical image data and doctor-patient dialogue data. The cephalometric medical image dataset is expanded using deep learning algorithms and the Steiner analysis method to generate diagnostic reports. Then, the vicuna-7b model is fine-tuned using the doctor-patient dialogue data. Finally, both the MiniGPT4 and VisualGLM models are fine-tuned using the cephalometric medical image dataset and the generated diagnostic reports. This approach is of great significance for the development of customized visual question answering systems and provides strong support for the practical application of large language models in the medical field.

With the CephGPT-4 model, doctors and orthodontic specialists can perform orthodontic measurement diagnostics more efficiently, reducing the workload and human errors associated with manual measurements. The model can also provide fast, consistent, and objective measurement results, which help guide the development and evaluation of orthodontic treatment plans, thereby reducing the barrier to use.

The innovation of this study lies in several aspects. Firstly, the proposed model is the first large language model applied in the field of orthodontics, which is pioneering and improves the time-consuming and labor-intensive issues in cephalometric analysis, enhancing the efficiency of treatment and alleviating the imbalance in medical institutions. Secondly, a multimodal orthodontic medical dataset is constructed, laying the foundation for future applications and developments of large language models in the field of orthodontics. Lastly, this study fine-tunes both the MiniGPT4 and VisualGLM-6B models and makes a comparison between the two models. With the development of deep learning techniques and large language models, future research can further explore other fine-tuning strategies and model architectures to further enhance the performance of visual question answering tasks. CephGPT-4 has the potential to be a valuable assistant in dental malocclusion correction. It combines the capabilities of natural language processing and computer vision to provide accurate diagnosis and treatment recommendations through the analysis of dental imaging data. Through communication with dentists and patients, it can provide personalized treatment plans, offering better dental health and aesthetics to patients.

## 6 Conclusion

The application of the CephGPT-4 model can not only improve the efficiency and accuracy of orthodontic measurements but also contribute to the integration of the medical field with natural language processing, promoting the development of medical image analysis and diagnostic techniques. However, the performance and reliability of the model need to be rigorously validated and evaluated to ensure reliable application in clinical practice.

## 7 Limitations and Issues

Due to the limited availability of publicly available datasets for cephalometric medical images, the created multimodal dataset in this study is relatively small, and the model training may not be comprehensive enough, which may affect the accuracy of the results. This study only uses datasets based on lateral cephalograms for fine-tuning, further work is needed to supplement and improve the work of large-scale cephalometric measurement models.

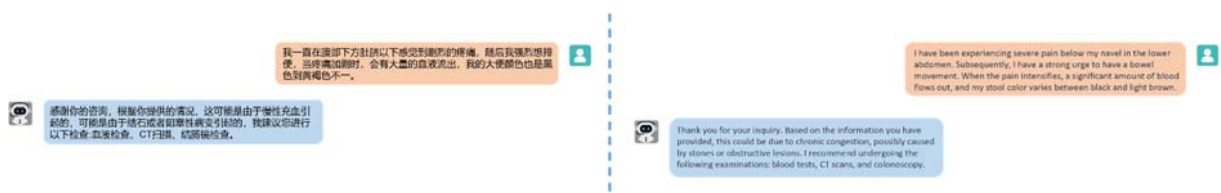

Figure 2 Chinese-English Dialogue Results after Fine-tuning Vicuna-7B on the MD-QA Dataset

These dialogue results demonstrate the interaction between the patient and the doctor in addressing common orthodontic-related questions. The Vicuna-7B model, after fine-tuning on the MD-QA dataset, effectively provides accurate translations and helpful responses to patient inquiries regarding orthodontic treatment.

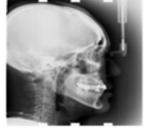

Figure 3 Comparison of Orthodontic Medical Image Question-Answering Performance Before and After Fine-tuning with Different Models

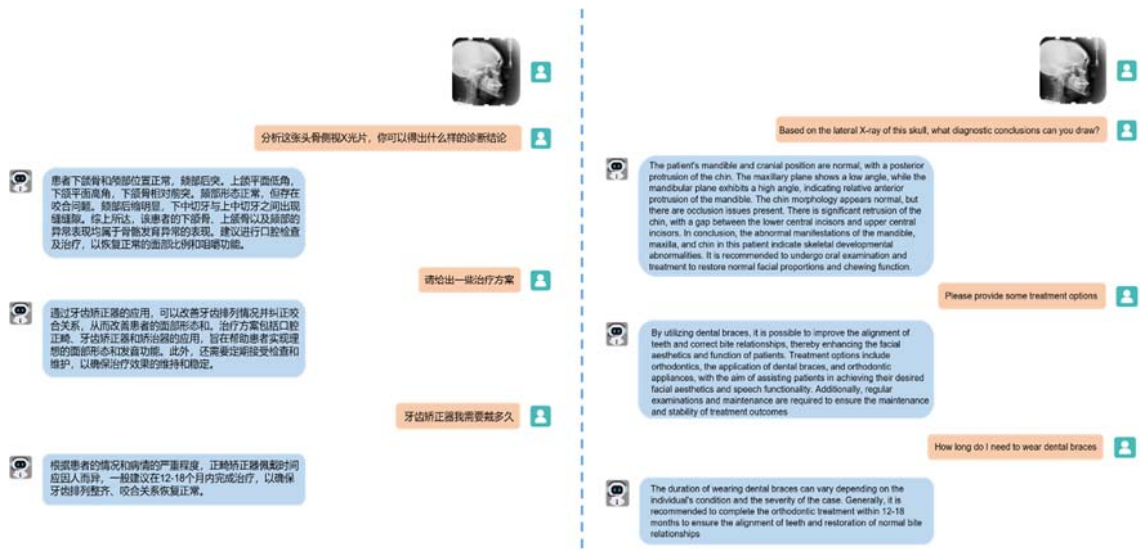

Figure 4 Diagnosis generated by CephGPT-4. The left side of Fig 3 is the Chinese result and the right side of fig3. is the Chinese result

Fig.4 presents the diagnosis generated by CephGPT-4. On the left side of the figure, you can find the diagnosis in Chinese, while on the right side, there is another diagnosis also written in Chinese. The figure showcases the generated diagnoses, providing information and insights based on the analysis performed by the CephGPT-4 model.

| Xray image | Lora | P-tuning |
|---|---|---|
|  | Based on the patient's lateral view X-ray of the skull, the maxilla appears to be in a normal position relative to the cranium, and the mandible shows a normal chin position with even angulation in the lower plane. However, the patient presents a skeletal Class II malocclusion, characterized by retrognathic mandible and protrusion of the maxillary incisors. Therefore, a diagnosis of dental developmental abnormalities can be confirmed, and orthodontic treatment is recommended to correct the skeletal deformity. | This lateral view X-ray of the skull shows evident fractures or injuries in the cranial, occipital, and parietal bones. The teeth are also affected to some extent. This may be the result of a head impact or other external trauma. However, further examination is required to determine the specific cause and develop a treatment plan |
|  | According to the patient's lateral view X-ray of the skull, the mandible exhibits a retrognathic position relative to the cranial base, while the maxillary incisors are in a normal position and there is a posterior concavity behind the mandibular incisors. The chin plane position is also normal, but there is anterior protrusion of the chin. Overall, the position of the mandible and maxilla is normal, but there is anterior protrusion of the chin. In summary, the patient has a skeletal Class II malocclusion and requires appropriate treatment. It is recommended to undergo orthodontic surgery to correct the mandibular angle, improve facial aesthetics, and restore proper occlusion, thus promoting oral health and aesthetics. | This lateral view X-ray of the skull shows the presence of cervical spine fracture, cervical disc protrusion, and spinal cord injury in the patient. These abnormalities may indicate the presence of cervical spondylosis or nerve damage, which can potentially lead to neck pain, mobility impairments, and other symptoms. Treatment options include rest, physical therapy, medication, and surgery, among others. If the condition is severe, it is advisable to seek advice and support from a healthcare professional to obtain the best treatment plan for symptom relief and recovery |

Figure 5, Dialogue results after fine-tuning by Lora and P-tuning

The left side of Fig.5 represents the dialogue results after fine-tuning with Lora, while the right side showcases the dialogue results after fine-tuning with P-tuning. By employing these two different fine-tuning techniques, a comparison is drawn to evaluate their respective impacts on the quality of the generated dialogue.